\documentclass[twocolumn,10pt,cleanfoot,cleanhead]{asme2e}

\def\BibTeX{{\rm B\kern-.05em{\sc i\kern-.025em b}\kern-.08em
    T\kern-.1667em\lower.7ex\hbox{E}\kern-.125emX}}
    
\usepackage{graphicx}
\usepackage{amsmath}

\title{Pipe Climbing Robot}

\author{Abdul Jalal, Ravi Kant, Arjun Kumar, V. Kumar}

\begin{document}

\maketitle    

\begin{abstract}
{\it This paper presents the plan of an in-pipe climbing robot that works utilizing a novel Three-Output Open Differential(3-OOD) component to navigate complex organizations of lines. Customary wheeled/followed in-pipe climbing robots are inclined to slip and haul while navigating in pipe twists. The 3-OOD component helps in accomplishing the original aftereffect of wiping out slip and drag in the robot tracks during movement. The proposed differential understands the practical capacities of the customary two-yield differential, which is accomplished the initial time for a differential with three results. The 3-OOD component precisely tweaks the track rates of the robot in light of the powers applied on each track inside the line organization, by wiping out the requirement for any dynamic control. The recreation of the robot crossing in the line network in various directions and in pipe-twists without slip shows the proposed plan's adequacy.}
\end{abstract}

\section*{INTRODUCTION}

Pipe networks are inescapable, basically used to ship fluids and gases in enterprises and metropolitan urban communities. Most frequently, the lines are covered to agree with the security rules and to stay away from hazards. This makes review and upkeep of lines truly challenging. Covered lines are exceptionally inclined to obstructing, consumption, scale arrangement, and break commencement, bringing about releases or harms that might prompt disastrous episodes.

Various Inspection Robots \cite{han2016analysis} were proposed in the past to lead ordinary preventive examinations to keep away from mishaps. Divider squeezed IPIRs with single, different and half breed systems \cite{roslin2012review}, Pipe Inspection Gauges (PIGs) \cite{okamoto1999autonomous}, actively controlled IPIRs with articulated joints and differential drive units \cite{ryew2000pipe} were additionally widely considered. Moreover, bio-motivated robots with crawler, inchworm, strolling components \cite{choi2007pipe}, and screw-drive\cite{kakogawa2013development} mechanisms were likewise demonstrated to be appropriate for various necessities. In any case, the majority of them utilize dynamic controlling techniques to guide and move inside the line. Reliance upon the robot's direction inside the line added to the difficulties, likewise leaving the robot powerless against slip on the off chance that foothold control techniques are not involved. The Theseus \cite{hirose1999design}, PipeTron \cite{debenest2014pipetron}, and PIRATE \cite{dertien2014design} robot series utilize separate sections for driving and driven modules that are interconnected by various linkage types. Each fragments adjust or change the direction for arranging turns. Also, hearty dynamic controlling makes such robots dependent on sensor information and weighty calculation.

Pipe climbing robots with three balanced modules are more steady and give better portability. Prior proposed secluded line climbers \cite{vadapalli2019modular,suryavanshi2020omnidirectional} have used three driving tracks arranged symmetrically to each other, similar to MRINSPECT series of robots \cite{roh2008modularized,roh2001actively,yang2014novel}. In such robots, to effectively control the three tracks, their speeds were pre-characterized for the line twists. This represented a constraint for the robot to arrange pipe-twists just at a specific direction comparing to the pre-characterized speeds \cite{vadapalli2019modular,suryavanshi2020omnidirectional,roh2008modularized,roh2001actively,yang2014novel}. In genuine applications, the robot's direction change assuming it encounters slip in the tracks during movement. This restriction can be addressed by utilizing an inactively worked differential component to control the robot. MRINSPECT-VI \cite{kim2016novel,kim2013pipe} utilizes a multi-pivotal differential stuff component to control the rates of the three modules. Notwithstanding, for the division of the driving force and speed to the three modules, the format of the differential is utilized. This methodology made the principal yield (Z) to pivot quicker than the other two results (X and Y), making yield Z effortlessly impacted by slip \cite{kim2016novel}. This is caused in light of the fact that the results of the differential doesn't impart comparable energy to the information. Other recently proposed answers for Three-output differentials (3-OD's) \cite{kota1997systematic,ospina2020sensorless} also followed a similar layout.

'Three-Output Open Differential' kills the referenced restriction by acknowledging identical result to enter active relations \cite{vadapalli2021design}. In the conceived format every one of the three results are similarly impacted by the info. This contributes for the robot to wipe out slip and drag in any direction of the robot during its movement. Moreover, the differential component in the line climber upgrades the convenience by decreasing the reliance on the dynamic controls to move through the line organizations.

The 3-OOD mechanism precisely moves force and speed from a solitary contribution to the robot's three tracks through complex stuff trains, in light of the heaps experienced by each track independently. 

\section{Design of the Modular Pipe Climber-III}

\subsection{Structure of the Robot}

The CAD model of the proposed robot, Modular Pipe Climber-III. An unpredictable nonagon focal frame of the robot houses three modules circumferentially $120^\circ$ separated from one another. The differential situated inside the focal case of the robot drives the three tracks by its driving sprockets through. slant outfits, that are associated with yields $(O_{1-3})$ from the 3-OOD. The itemized perspective on the 3-OOD instrument and is definite in the ensuing subsection.

Every module houses a track and has openings for the four linkages to slide. The modules are pushed radially outwards with the assistance of direct springs mounted on the linkages (or shafts) i.e., the springs between the modules and the robot body. The plug is a reference structure that confines the movement of the modules past admissible cutoff points. At the point when the robot is sent inside the line, the spring-stacked tracks go through aloof avoidance and presses against the line's inward dividers. It gives the vital foothold to the robot to move. Every module in the robot can likewise pack unevenly, point by point in segment III.

\vspace{-0.15in}
\begin{figure}[ht!]
\centering
\includegraphics[width=3.4in]{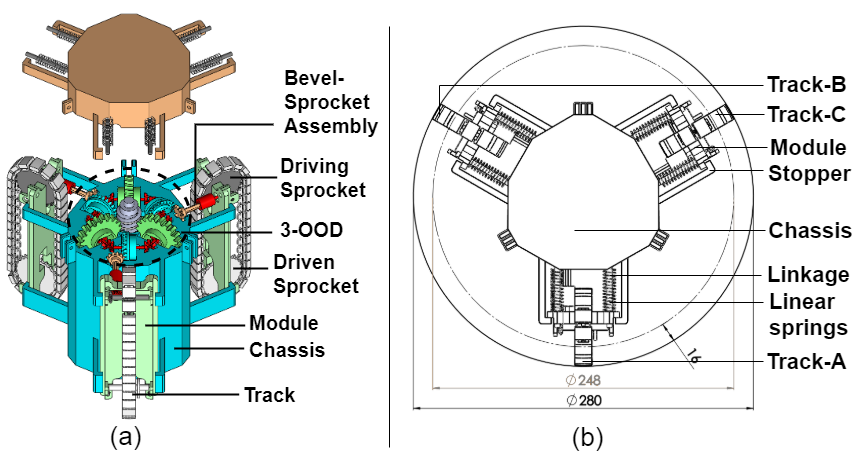}
\vspace{-0.15in}
\caption{\footnotesize (a) Robot (b) Robot's top-view}
\label{1}
\vspace{-0.15in}
\end{figure}
 
 divider squeezing component of the robot. The tracks, obliged on the modules, are squeezed against the internal dividers of the line which give the important footing to the robot to move. There is a driving sprocket that changes over the rotational movement from the result of 3-ODD to translatory movement for the robot. The divider squeezing system comprises of 4 direct linkages for every one of the three modules. The modules have openings through which the straight linkages (or shafts) pass. They permit the movement of the modules in the spiral headings from the pivot of the robot.

\subsection{Three-Output Open Differential (3-ODD)}

The Three-Output Open Differential (3-OOD) is the essential constituent of the proposed robot. The 3-OOD instrument contains a solitary information $(U)$, three two-yield open differentials $(2-OD_{1-3})$, three two-input open differentials $(2-ID_{1-3})$ and three results $(O_{1-3})$. The differential's feedback $(U)$ is situated at the focal pivot of the robot body. The three $2-OD_{1-3}$ are organized evenly around the info $(U)$, with a point of $120{^\circ}$ between any two. The $2-ID_{1-3}$ are fitted radially in the middle $2-OD_{1-3}$. The single result of $2-ID_{1-3}$ structure the three results $(O_{1-3})$ of the 3-OOD. $2-OD_{1-3}$ involves gear components, for example, ring gears $(R_{1-3})$, slant gears $(B_{1-6})$ and side cog wheels $(S_{1-6})$, while $2-ID_{1-3}$ incorporate ring gears $(R_{4-6})$, slant gears $(B_{7-12})$ and side pinion wheels $(S_{7-12})$. The side cog wheels $(S_{1-6})$ of $2-OD_{1-3}$ is coincided with their nearby side cog wheels $(S_{7-12})$ of $2-ID_{1-3}$, to move the force and speed from $2-OD_{1-3}$ to it's neighboring $2-ID_{1-3}$. The information $(U)$ of the worm gear give movement to $2-OD_{1-3}$ simuntaneously. Every two-yield differential $(2-OD_{1-3})$ then exchange got movement to its adjoining two-input differentials $(2-ID_{1-3})$, contingent upon the heap conditions experienced by its separate side pinion wheels $(S_{1-6})$. The movement got by the side pinion wheels $(S_{7-12})$ of $2-ID_{1-3}$ further makes an interpretation of them to the three results ($O_{1-3})$. The six differentials ($2-OD_{1-3}$ and $2-ID_{1-3})$ work together to interpret movement from the contribution to the three results ($O_{1-3})$.

At the point when $O_{1-3}$ experience various burdens, the side pinion wheels ($S_{7-12})$ in $2-ID_{1-3}$ exchanges various burdens to the side cog wheels $(S_{1-6})$ of $2-OD_{1-3}$. Under this condition, $2-OD_{1-3}$ makes an interpretation of differential speed to its neighboring $2-ID_{1-3}$. At the point when $O_{1-3}$ experience equivalent burden or no heap, every one of the side pinion wheels experience a similar burden causing them to pivot at a similar speed and force. As laid out, every one of the results ($O_1$, $O_2$, $O_3$) share equivalent energy with the info. Furthermore, the results likewise share indistinguishable energy with one another. This outcomes in the difference in loads in one of the results forcing an equivalent impact on the other two results that are undisturbed. The results work with equivalent velocities when there is no heap or equivalent burden following up on every one of the results. The 3-OOD component works its results with differential speed when the results are under changed burdens. At the point when one of the results is working at an alternate speed while the other two results are encountering similar burden, then, at that point, the two results with similar burdens will work with equivalent velocities. The 3-OOD component intended for the Modular Pipe Climber III, propels its the three tracks with identical paces while moving inside a straight line. In any case, while moving inside the line twists, the differential adjusts the track speed of the robot with the end goal that the track venturing to every part of the more extended distance turns quicker than the track venturing to every part of the more limited distance. Refer to \cite{kumar2021design} for more figures, graphs and information.

\section{Kinematics and Dynamics}

\subsection{Kinematics and Dynamics of the 3-OOD mechanism}

The kinematic plot shows the availability of the connections and joints of the 3-OOD instrument. The Three-Output Open Differential's kinematic and dynamic conditions are planned utilizing the bond diagram model. The info precise speed ($\omega_u$) from the engine is similarly conveyed to the three ring cog wheels of the two-yield differentials as $\omega_{r1}$, $\omega_{r2}$ and $\omega_{r3}$. They turn at equivalent precise speeds and with equivalent force for example $1/k$ times $\omega_u$ and $k/3$ occasions $\tau_u$, where $1/k$ is the stuff proportion of the contribution to the ring gears ($1/k$ = $1/20$). Besides, a two-yield differential directs that the precise speed of its ring gear is the normal every time of the rakish speeds of its two side pinion wheels. These two side cog wheels can pivot at various rates while keeping up with equivalent force \cite{deur2010modeling}.

\begin{equation}
\omega=\frac{k(\omega _{01}+\omega _{02})}{2}=\frac{k(\omega _0{3}+\omega _{04})}{2}=\frac{k(\omega _{05}+\omega _{06})}{2},
\label{1}
\end{equation}

From the bond chart model, we can derive that the three ring gears ($R_{1-3}$) pivots at equivalent precise speeds and with equivalent force for example $1/k$ times the info rakish speed, $\omega_u$ and $k/3$ occasions the information force, $\tau_i$, where $1/k$ is the stuff proportion of the contribution to the ring gears ($1/k$ = $1/20$). Additionally, in a two-yield differential, the precise speed of its ring gear is the normal all of the time of the rakish speeds of its two side pinion wheels. These two side cog wheels can turn at various paces while keeping up with equivalent force \cite{vadapalli2021design}. 

The rakish speeds $\omega_{r1}$, $\omega_{r2}$ and $\omega_{r3}$ are then meant their side cog wheels. Side cog wheels $(S_1$, $S_7$), $(S_3$, $S_8$) and $(S_5$, $S_{11}$) are associated with the end goal that they don't have any overall movement between the pinion wheels of a similar pair, i.e., precise speeds of side cog wheels of a similar pair is generally equivalent $(\omega_1$$=$$\omega_7)$, $(\omega_3$$=$$\omega_8)$ and $(\omega_5$$=$$\omega_{11})$. Subbing in (\ref{1})

Essentially, the result precise speeds $\omega_{O1}$, $\omega_{O2}$, $\omega_{O3}$ are gotten from the speeds $\omega_{r4}$, $\omega_{r5}$, $\omega_{r6}$ of the particular ring gears $R_4$, $R_5$, and $R_6$. The result to side stuff connection is compared to accomplish the rakish speed condition for the contribution to the results. The angular velocities of the ring gears to the side gears, we attain a relation between input and side gears. Similarly, the output angular velocities $\omega_{O1}$,  $\omega_{O2}$,  $\omega_{O3}$ are calculated from the velocities $\omega_4$, $\omega_5$, $\omega_6$  of the respective ring gears $R_4$, $R_5$, and $R_6$ to get an output to side gear relation. Equating both the relations, we attain the angular velocity equation for the input to the outputs.

where $\omega_u$ is the precise speed of the info $(U)$, $j$ is the stuff proportion of the ring gears $(R_{4-6})$ to the results ($j$ = 2:1), while $\omega_{O1}$, $\omega_{O2}$ and $\omega_{O3}$ are rakish speeds of the result. $\omega_1$, $\omega_2$, $\omega_3$, $\omega_4$, $\omega_5$ and $\omega_6$ are the individual precise speeds of the side cog wheels $(S_1$, $S_2$, $S_3$, $S_4$, $S_5$, $S_6)$. Subsequently, the result precise speeds ($\omega_{O1}$, $\omega_{O2}$, $\omega_{O3}$) are subject to the info rakish speed ($\omega_u$) and the side stuff yields ($\omega_2$, $\omega_4$), ($\omega_3$, $\omega_5$) and ($\omega_1$, $\omega_6$). In the mean time, the forces $\tau_{R4}$, $\tau_{R5}$ and $\tau_{R6}$ of the ring gears $(R_4$, $R_5$ and $R_6)$ is the amount of the forces of their comparing side pinion wheels $S_7$, $S_8$, $S_9$, $S_{10}$, $S_{11}$ and $S_{12}$. Like precise speed connection for contribution to yields, by comparing the ring pinion wheels to side pinion wheels connection with the result to the ring gear connection, a connection between yield forces ($\tau_{O1}$, $\tau_{O2}$, $\tau_{O3}$) to the information force ($\tau_{u}$) is acquired.

\begin{equation}
\begin{split}
& \tau_{O01}=\frac{k(\tau_u)}{3j}-\frac{(I_{01}\Dot{\omega}_{07}+I_{03}\Dot{\omega}_{08})}{j}, \ \hspace{0.25in}  \tau_{O02}=\frac{k(\tau_u)}{3j}\hspace{0.02in}- \\ &\frac{(I_{04}\Dot{\omega}_{09}+I_{06}\Dot{\omega}_{010})}{j}, \ \hspace{0.25in}  \tau_{O03}=\frac{k(\tau_u)}{3j}-\frac{(I_{02}\Dot{\omega} _{012}+I_{05}\Dot{\omega}_{011})}{j}
\label{2}
\end{split}
\end{equation}

where $I_1$,$I_2$,$I_3$,$I_4$,$I_5$,$I_6$ are the inactivity shown by the side pinion wheels, $\Dot{\omega}_7$, $\Dot{\omega}_8$, $\Dot{\omega}_9$, $\Dot{\omega}_{10}$, $\Dot{\omega}_{11}$, $\Dot{\omega}_{12}$ are the precise speed increase of the separate side pinion wheels and $\tau_u$ is the information force. The 3-OOD instrument has three levels of opportunity in its result. Condition (\ref{2}) approves that the conduct of each result is affected by the info ($U$) just as the other two results. \textbf{\small Equal speeds and torque:\normalsize} Indistinguishable side pinion wheels $(S_{7-12})$ display equivalent idleness $(I_1=I_2=I_3=I_4=I_5=I_6)$. The side pinion wheels $(S_{7-12})$ work with equivalent rakish speed $(\omega _n= \frac{\omega _i}{k}$, where $n$ goes from $1$ to $12)$ and precise speed increase when every one of the three results $(O_{1-3})$ experience equivalent burdens. Subbing these relations in \eqref{2} we get,

\begin{equation}
{\bf\omega _{O01}}={\bf\omega _{O02}}={\bf\omega _{O03}}=\frac{j(\omega _u)}{k}
\label{3}
\end{equation}

\begin{equation}
{\bf\tau _{O01}}={\bf\tau _{O02}}={\bf\tau _{O03}}=\frac{k(\tau_u)}{3j} - \frac{2(I_1\dot{\omega}_1)}{j}
\label{4}
\end{equation}

Conditions \eqref{3} and \eqref{4} presents the curiosity of the differential to give equivalent movement attributes in every one of the three results that encounters equivalent burdens or when left unconstrained. However, when the stuff parts experience an obstruction across an intersection, the rakish speed and the force changes relying upon the outer resistive power. The result speed of the differential results are equivalent to the speed of the driving module sprockets.Therefore, the result speeds ${\omega_{O1}}$, ${\omega_{O2}}$ and ${\omega_{O3}}$ of the differential are converted into track speeds $v_{tA}$, $v_{tB}$ and $v_{tC}$. The info speed for the robot is 120 rpm, consequently making an interpretation of 12 rpm to the results under equivalent stacking conditions. The sprocket breadth is steady ($D_s$= 80 mm) for every one of the three tracks.

Ho Moon Kim et al. \cite{kim2013pipe}, in their paper recommended a technique for working out the particular speeds of the three tracks inside pipe twists. Accepting that the robot enters the pipeline with the design displayed is the point of the line twist, R is the sweep of ebb and flow of the line curve and r is the range of the line. The speed of the track An is gotten from

\begin{equation}
v_{A} = v(\frac{R-r\cos({\mu})}{R})
\label{eqn01}
\end{equation}

Likewise, the speeds $v_{tB}$ and $v_{tC}$ for their individual tracks B and C are gotten. The robot is embedded at various direction of the modules as for OD. The adjusted paces for the tracks are determined for twist pipes in directions $\mu$ = $0^\circ$, $\mu$ = $30^\circ$, $\mu$ = $60^\circ$.

\subsection{Asymmetrical Compression}

The straight springs in the module gives robot the adaptability to arrange twists without any problem. The greatest pressure conceivable in every module is $16 mm$. There are extra resiliences in the module openings, with the goal that topsy-turvy pressure is conceivable. This assists the robot with defeating hindrances and anomalies in the line network that it might look in true applications. The front finish of the module is compacted totally though the backside is in its greatest broadened state conceivable inside the line. The most extreme lopsided pressure permitted in a solitary module of the robot. Consequently, $\phi$ is the greatest point the module can pack unevenly. Refer to \cite{kumar2021design} for more figures, graphs and information.

\vspace{-0.1in}
\begin{figure}[ht!]
\vspace{-0.1in}
\centering
\includegraphics[width=3in]{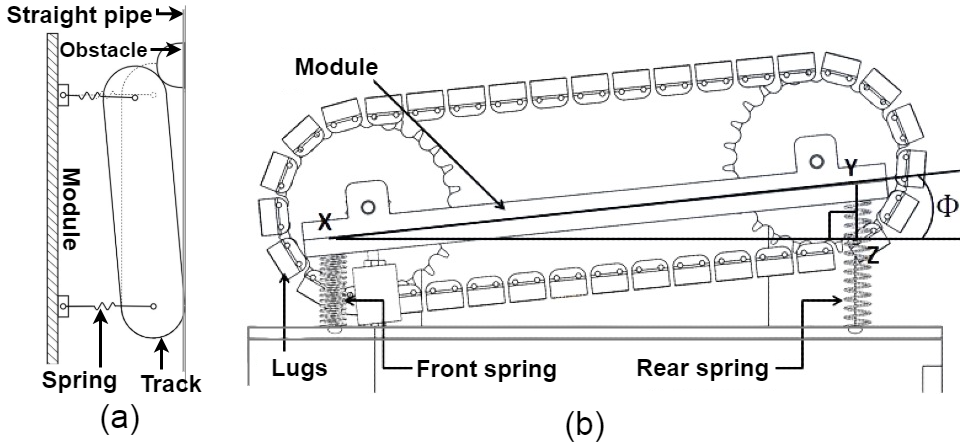}
\vspace{-0.12in}
\caption{\footnotesize (a) Robot overcoming obstacle (b) Asymmetrical module compression}
\label{asym(1)}
\vspace{-0.1in}
\end{figure}

\section{Dynamic Simulations of the Pipe Climber }

Recreations were led to investigate and approve the movement capacities of the robot in different line organizations. A similar will give us more experiences into the elements and conduct of the created Modular Pipe Climber-III in genuine testing conditions. Henceforth, multi-body dynamic reproductions was acted in MSC Adams by changing over the plan boundaries into an improved on reenactment model. To diminish the quantity of moving parts in the model and to diminish the computational burden, the tracks were rearranged into roller wheels. Every module houses three roller wheels in the improved on model. Along these lines, the contact fix given by the tracks to the line dividers are decreased from 10 contact locales to 3 contact areas for every module. The reenactment boundaries like the track speeds ($v_{tA}$, $v_{tB}$, and $v_{tC}$) and the module pressure for each track A,B and C were examined. Reproductions were performed by embedding the robot in three distinct directions of the module ($\mu$ = 0${^\circ}$, $\mu$ = 30${^\circ}$, $\mu$ = 60${^\circ}$) in both the straight lines and line twists. The robot is reproduced inside a line network planned by ASME B16.9 standard NPS 11 and timetable 40. The recreations were directed for four experiment situations in the line network comprising of Vertical area, Elbow segment (90${^\circ}$ curve), Horizontal segment and the U-segment (180${^\circ}$) for various directions ($\mu$ = 0${^\circ}$, $\mu$ = 30${^\circ}$, $\mu$ = 60${^\circ}$) of the robot. The absolute distance of the line structure is $D_{pipe}$ = 3,023.49mm. The distance went by the robot ($D_{R}$) in pipe is determined from focal point of the robot body and the track's singular distance voyaged is determined from the middle roller wheel mounted in every modules. In this way, we get the absolute robot's way, by taking away the robot's length from $D_{pipe}$ (i.e., $D_{R}$ =$D_{pipe}$ - $(L_R)$ = 2,823.49mm, where $L_R$= 200mm).The robot's way in vertical climbing and the last flat area is estimated by deducting $L_R/2$ from their individual segment length. The info ($U$) of the 3-OOD is given a consistent rakish speed of 120rpm ($\omega_u$= 120rpm) and movement of the robot including the track speeds ($v_{tA}$, $v_{tB}$, and $v_{tC}$) are considered in the reenactment.

\subsection{Vertical section and Horizontal section}

In the upward segment and level segment, the robot follows a straight way. Consequently, the tracks experience equivalent burdens on every one of the three modules in both the experiments. Therefore, the differential gives equivalent speeds to every one of the three tracks, comparable to the robot's normal speed $v_R$. The noticed track speeds in the recreation for the direction $\mu$ = 0$^\circ$, is $v_R$= $v_{tA}$= $v_{tB}$= $v_{tC}$= 50.03 mm/sec. Likewise, for $\mu$ = 30$^\circ$, the speeds are $v_R$= $v_{tA}$= $v_{tB}$= $v_{tC}$= 50.22 mm/sec and for $\mu$ = 60$^\circ$, the speeds are $v_R$= $v_{tA}$= $v_{tB}$= $v_{tC}$= 51.36 mm/sec. In this manner, every one of the qualities compare to the hypothetical outcomes with a flat out rate mistake (APE) lesser than 2.2\%. This mistake evaluates the genuine measure of deviation from the hypothetical worth \cite{armstrong1992error}. To arrange an underlying length of 550 mm ($D_{R}$ = 550 - $(L_R/2)$= 450 mm) in vertical climbing, the robot requires 0 to 9 seconds. the robot's capacity to climb the line upward against gravity. Beginning from 24 seconds till 31 seconds, the robot moves a distance of 350 mm in the principal even area. In the more modest flat segment of distance 150mm ($D_{R}$ = 150 - $(L_R/2)$= 50 mm), the robot moves from 59 to 60 seconds.

\vspace{-0.1in}
\subsection{Elbow section and U-section}

In the elbow area ($90^\circ$ twist) and U-segment ($180^\circ$ twist), the robot moves at a steady distance to the focal point of arch of the line. The 3-OOD system adjusts the result paces of the tracks $v_{tA}$, $v_{tB}$ and $v_{tC}$ as indicated by the separation from the focal point of shape of the line. In each of the three directions of the robot, the external module tracks turns quicker to travel a more extended distance, while the inward module tracks pivots more slow to venture to every part of the most limited distance than the span of curve of the line twist. In pipe twists, the recreation speeds of each track is arrived at the midpoint of seperately to rough the noticed track speeds without changes. The approximated speeds of each track is then contrasted with their particular hypothetical speeds with track down the outright rate blunder (APE). For the direction $\mu$ = 0$^\circ$, the external modules (B and C) move at a normal speed of (58.7 mm/sec and 57.8 mm/sec), while the inward module (A) move at a normal speed of 33.62mm/sec. These qualities relate to the hypothetical qualities $v_{tB}$ = $v_{tC}$ = 58.51 mm/sec and $v_{tA}$ = 33.69mm/sec with an APE of 1.2\%. Additionally, the track speeds $v_{tA}$, $v_{tB}$ and $v_{tC}$ for $\mu$ = 30$^\circ$, coordinates with the normal worth of the reproduction results ($v_{tC}$= 50.3 mm/sec, $v_{tB}$= 63.8 mm/sec, $v_{tA}$= 37.3 mm/sec) with an APE of 3.8\%. Similarly, the track speed an incentive for $\mu$ = 60$^\circ$, compare to the reproduction results ($v_{tB}$= 68.5 mm/sec, $v_{tA}$= 40.2 mm/sec and $v_{tC}$= 41.3 mm/sec) with an APE of 2.5\%. In every directions of both the straight and twist areas, the mistake esteem is extremely negligible and they can be credited because of outside factors in true conditions like erosion. In this manner, negligible speed changes happens in the reproduction plot. From 9 to 24 seconds, the robot arranges the elbow area (90$^\circ$ twist) of distance 657.83 mm, while it requires 31 to 59 seconds to travel 1315.66 mm in the U-segment (180$^\circ$ curve). The robot's capacity to navigate in pipe-twists. for $\mu$ = 0$^\circ$ shows that the external modules (B and C) move at the speed scope of (52-60mm/sec) while the inward module (A) move in a speed scope of (30-37.25mm/sec). These qualities exist in the hypothetical qualities $v_{tB}$ = $v_{tC}$ = 58.51 mm/sec and $v_{tA}$ = 33.69mm/sec. The track speeds $v_{tA}$, $v_{tB}$ and $v_{tC}$ for $\mu$ = 30$^\circ$, coordinates with the reenactment results ($v_{tC}$= 49-52mm/sec, $v_{tB}$= 60-75mm/sec, $v_{tA}$= 32.5-40mm/sec). Also, the track speeds an incentive for $\mu$ = 60$^\circ$, exist in the reenactment results($v_{tB}$= 62-75mm/sec and $v_{tA}$= $v_{tC}$= 32.5-48). The speed range referenced from the recreation is taken from the base to the most extreme pinnacles of speeds for each plots. Refer to \cite{kumar2021design} for more figures, graphs and information.

The reproduction results for the track speeds $v_{tA}$, $v_{tB}$, $v_{tC}$ and the robot $v_{R}$ in various directions ($\mu$ = 0$^\circ$, $\mu$ = 30$^\circ$, $\mu$ = 60$^\circ$), coordinates with the hypothetical outcomes got in area III. It is seen from the reenactment that in 60 seconds, the robot navigates all through the line network consistently at the embedded direction. The outcome relate to the hypothetical computation ($D_{R}$/$v$= 3016.49/50.24 = 60.04 sec). This approves that the Three-Output Open Differential disposes of slip and drag in the tracks of the line climber in all directions with no movement misfortunes. In the reproduction, the robot is seen with no slip and drag in all directions, which further effects in decreased pressure impact on the robot and expanded movement perfection. \textbf{\small Radial flexibility:\normalsize} The track in the modules clasp to the internal mass of the line to give footing during movement. The springs are at first pre-stacked by a pressure of 1.25 mm in each of the three modules similarly when embedded in the upward line area. In straight lines, the robot moves at the underlying pre-stacked spring length. The deformity length increments by 1.5 mm for the internal and the external modules when the robot is moving close to elbow area and U-segment. This disfigurement clarifies the outspread adaptability took into consideration the modules to move through the evolving cross-segment of the line breadth in the curves during movement.

\section{Conclusion and Future works}

\textbf{\small Eliminating slip and drag:\normalsize}
The Modular Pipe Climber-III robot is given the clever Three-Output Open Differential to control the robot precisely with no dynamic controls. The differential has an identical result to include energy, whose presentation is totally similar to the usefulness of the customary two result differential. The reproduction results approve effective crossing of mind boggling pipe networks with curves of up to 180$^\circ$ in various directions without slip. Taking on the differential component in the robot accomplishes the clever aftereffect of wiping out the slip and drag in all directions of the robot during the movement. At the present, we are fostering a model to perform investigates the proposed plan \cite{vadapalli2021modular}.

\bibliographystyle{asmems4}
%
% ...

\bibliography{asme2e}

\end{document}